\documentclass[10pt,letterpaper]{article}
\usepackage[top=0.85in,left=2.75in,footskip=0.75in,marginparwidth=2in]{geometry}

\usepackage[utf8x]{inputenc}

\usepackage{cite}

\usepackage{nameref,hyperref}

\usepackage[right]{lineno}

\usepackage{microtype}
\DisableLigatures[f]{encoding = *, family = * }

\raggedright
\usepackage{changepage}

\usepackage[aboveskip=1pt,labelfont=bf,labelsep=period,singlelinecheck=off]{caption}

\makeatletter
\renewcommand{\@biblabel}[1]{\quad#1.}
\makeatother

\usepackage{lastpage,fancyhdr,graphicx}
\usepackage{epstopdf}
\fancyheadoffset[L]{2.25in}
\fancyfootoffset[L]{2.25in}

\usepackage{color}

\definecolor{Gray}{gray}{.25}

\usepackage{graphicx}

\usepackage{sidecap}

\usepackage{wrapfig}
\usepackage[pscoord]{eso-pic}
\usepackage[fulladjust]{marginnote}
\reversemarginpar

\begin{document}
\vspace*{0.35in}

\begin{flushleft}
{\Large
\textbf\newline{Automatic detection and decoding of honey bee waggle dances}
}
\newline
\\
Fernando Wario,
Benjamin Wild,
Raúl Rojas,
Tim Landgraf\textsuperscript{*}
\\
\bigskip
FB Mathematik und Informatik, Freie Universität Berlin , Berlin, Germany
\\
\bigskip
* tim.landgraf@fu-berlin.de

\end{flushleft}

\section*{Abstract}
The waggle dance is one of the most popular examples of animal communication. Forager bees direct their nestmates to profitable resources via a complex motor display. Essentially, the dance encodes the polar coordinates to the resource in the field. Unemployed foragers follow the dancer's movements and then search for the advertised spots in the field. Throughout the last decades, biologists have employed different techniques to measure key characteristics of the waggle dance and decode the information it conveys. Early techniques involved the use of protractors and stopwatches to measure the dance orientation and duration directly from the observation hive. Recent approaches employ digital video recordings and manual measurements on screen. However, manual approaches are very time-consuming. Most studies, therefore, regard only small numbers of animals in short periods of time. We have developed a system capable of automatically detecting, decoding and mapping communication dances in real-time. In this paper, we describe our recording setup, the image processing steps performed for dance detection and decoding and an algorithm to map dances to the field. The proposed system performs with a detection accuracy of 90.07\%. The decoded waggle orientation has an average error of -2.92° ($\pm$ 7.37° ), well within the range of human error. To evaluate and exemplify the system's performance, a group of bees was trained to an artificial feeder, and all dances in the colony were automatically detected, decoded and mapped. The system presented here is the first of this kind made publicly available, including source code and hardware specifications. We hope this will foster quantitative analyses of the honey bee waggle dance.

\section*{Introduction}
\begin{figure}[!ht]
\centering
\includegraphics[width=1.0\textwidth]{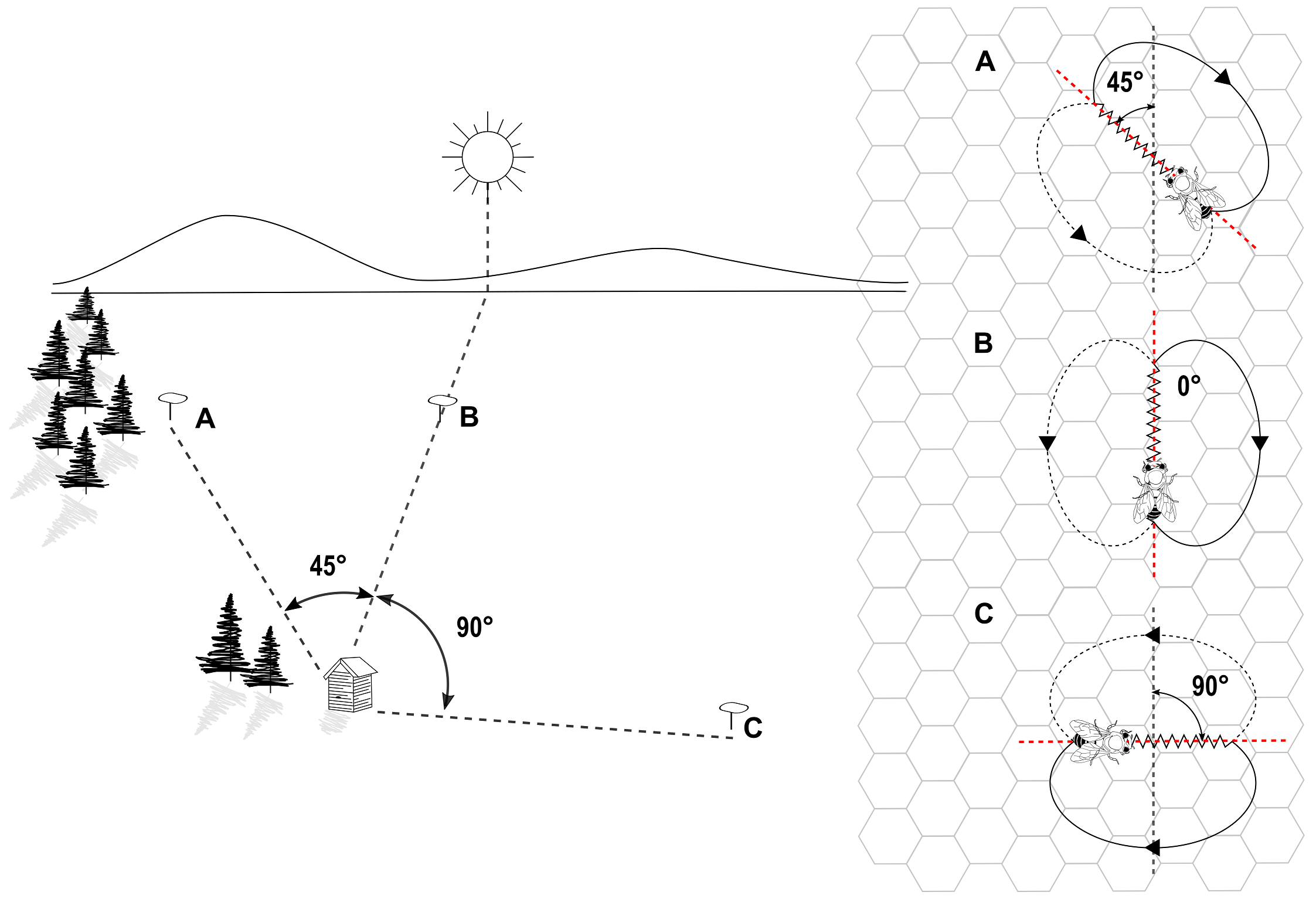}
\caption{{\bf Correlation between waggle dance parameters and locations on the field.}
On the left, three food sources in the field located at A) 45° counterclockwise B) 0° and C) 90° clockwise, with respect to the azimuth. On the right, their representation through waggle dance paths on the surface of a vertical honeycomb.}
\label{fig1}
\end{figure}

The honey bee waggle dance is one of the most popular communication systems in the animal world. Forager bees move in a stereotypic pattern on the honeycomb to share the location of valuable resources with their nestmates ~\cite{Frisch1965a,Seeley1995a,Gruter2008a}. Dances consist of waggle and return phases. During the waggle phase, the dancer vibrates her body from side to side while moving forward in a rather straight line on the vertical comb surface. Each waggle phase is followed by a return phase, during which the dancer circles back to the starting point of the waggle phase. Clockwise and counterclockwise return phases are alternated, such that the dancer describes a path resembling the figure eight ~\cite{Frisch1965a,Weidenmuller1999,Tanner2010,Landgraf2011a}. The average orientation of successive waggle runs with respect to gravity approximates the angle between the advertised resource and the solar azimuth as seen from the hive Fig~\ref{fig1}. The duration of the waggle run correlates with the distance between hive and resource ~\cite{Frisch1965a,Esch1996,Esch2001,Dacke2008}. The resource's profitability is encoded in the dance tempo: valuable resources are signaled with shorter return runs, yielding a higher waggle production rate ~\cite{Seeley2000}.

Unemployed foragers might become interested in a waggle dance, follow the dancer's movements and decode the information contained in the dance. Followers may then exit the hive to search for the communicated resource location in the field ~\cite{Seeley1995a,Biesmeijer2005,Riley2005a,Menzel2011,AlToufailia2013b}. Recruits that were able to locate the resource, once back in the hive, may also dance, thereby amplifying the collective foraging effort.

The study of the waggle dance as an abstract form of communication received great interest after it was first described by von Frisch ~\cite{VonFrisch1946}. Keeping bee colonies in special hives for observation is well-established. The complex dance behavior allows insights into many aspects of the honey bee biology and, even after seven decades, several research fields investigate the waggle dance communication system.

The dance essentially contains polar coordinates for a field location. Hence, waggle dances can be mapped back to the field ~\cite{Visscher1982}. The directional component relies on fixed reference systems such as gravity or the sun's azimuth and therefore is straightforward to compute from a dance observation. Honeybees integrate the optical flow they perceive along their foraging routes to gauge the distance they have flown ~\cite{Esch2001}. Several factors affect the amount of optical flow perceived, such as wind (bees fly closer to the ground in with strong headwinds ~\cite{Barron2006}) or the density of objects in the environment, such as vegetation or buildings. Honeybees calibrate their odometer to the environment before engaging in foraging activities. Hence, to convert waggle durations to feeder distances, our system requires calibration itself. To this end, bees must be trained to a number of sample locations with known distance. Assuming homogeneous object density in all directions, it may be sufficient to use a simple conversion factor obtained from the waggle durations observed in dances for a single feeding location.

This way, without tracking the foragers' flights, one can deduce the distribution of foragers in the environment by establishing the distribution of dance-communicated locations. Couvillon et al. ~\cite{Couvillon2014b} used this method to investigate how the decline in flower-rich areas affects honey bee foraging, while Balfour and Ratnieks ~\cite{Balfour2016} used it to find new opportunities for maximizing pollination of managed honey bee colonies. But mapping is not the only application of decoding bee dances. Theoretical biologists have studied the information content of the dance ~\cite{Haldane1954,Schurch2015} and the accuracy and precision with which bees represent spatial information through waggle dances ~\cite{DeMarco2008,Couvillon2012a}. Landgraf and co-workers tracked honey bee dances in video recordings to build a motion model for a dancing honey bee robot ~\cite{Landgraf2011a,Landgraf2012}. Studies on honey bee collective foraging also focus on the waggle dance ~\cite{Seeley1986}, including studies that model their collective foraging ~\cite{Camazine1991,DeVries1998a} and nest-site selection behavior ~\cite{Seeley2001,Passino2006,Reina2015}. In ~\cite{Wario2015} we automatically decoded waggle dances as part of an integrated solution for the automatic long-term tracking of activity inside the hive.

Different techniques have been used over time to decode waggle dances. During the first decades that followed von Frisch's discovery, most of the dances were analyzed in real time, directly from the observation hive with the help of protractors and stopwatches ~\cite{Frisch1965a,Seeley1995a}. Throughout the last decade, the use of digital video has become ubiquitous to extract the encoded information on their computers. Digital video allows researchers to analyze dances frame by frame and extract their characteristics either manually using the screen as a virtual observation hive ~\cite{Couvillon2012a}, or assisted by computer software ~\cite{DeMarco2008}. Although digital video recordings allow measurements with higher accuracy and precision, decoding communication dances continued to be a manual and time-consuming task.

Multiple automatic and semi-automatic solutions have been proposed to simplify and accelerate the dance decoding process. A first group of solutions focused on mapping the bees' trajectories via tracking algorithms ~\cite{Khan2004,Kimura2011a,Landgraf2007a}. These trajectories might then be analyzed to extract specific features such as waggle run orientation and duration, using either a generic classifier trained on bee dances (see ~\cite{Feldman2003,Kabra2012b,Oh2008}) or methods based on hand-crafted features such as the specific spectral composition of the trajectory in a short window ~\cite{Landgraf2011a}. Although a method has been described by Feldman and Balch ~\cite{Feldman2004} that could potentially be an automatic detector and decoder of dances ~\cite{Feldman2004}, its implementation has been limited to the automatic labeling of behaviors.

Here we propose a solution to detect and decode waggle dances automatically. Since all information known to be carried in the dances, can be inferred from the waggle run characteristics, our algorithm exclusively detects this portion of the dance directly from the video stream, avoiding a separate tracking stage. Our system detects 89.8\% of all waggle runs with a false positive rate of only 5\%. Compared to a human observer, the system extracts the waggle orientation with an average error of -2.92° ($\pm$ 7.37°) well within the range of human error.

\section*{Materials and Methods}

\subsection*{Hive and recording setup}
\begin{figure}[!ht]
\centering
\includegraphics[width=1.0\textwidth]{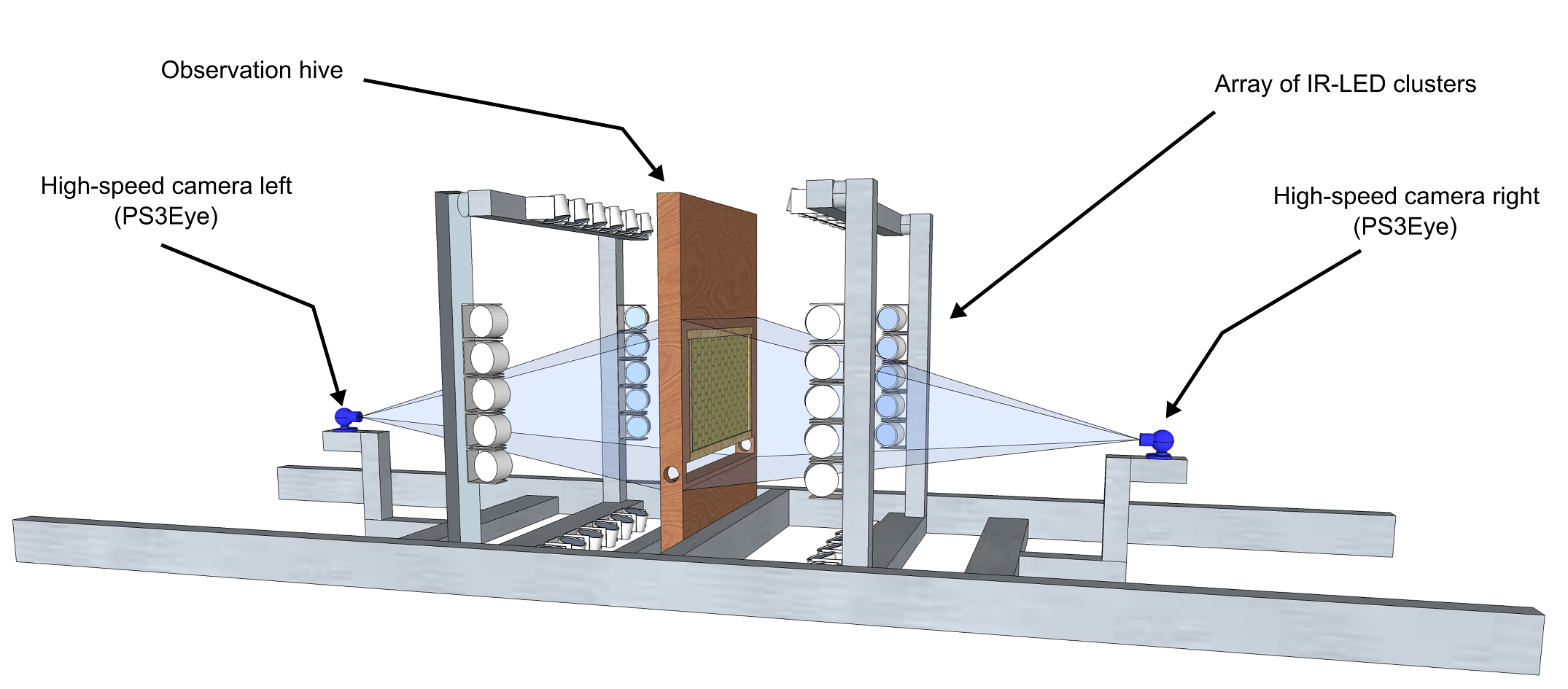}
\caption{{\bf Recording setup.}
A basic recording setup consists of an observation hive, an array of LED clusters for illumination and one webcam per side.}
\label{fig2}
\end{figure}

Our solution can be used either online with live streaming video or offline with recorded videos. The software requires a frame rate of approximately 100 Hz and a resolution of at least 1.5 pixels per millimeter. Thus QVGA resolution suffices to cover the whole surface of a ``Deutschnormal'' frame (370 mm x 210 mm). The four frame corners are used as a reference to rectify distortions caused by skewed viewing angles or camera rotations; If the frame corners are not captured by the camera, it is necessary to consider other reference points and their relative coordinates. A basic setup configuration is depicted in Fig~\ref{fig2} and might serve as a template for the interested researcher.

In our experiments, we worked with a small colony of 2000 bees (\textit{Apis mellifera carnica}). We used a one frame observation hive and one modified PS3eye camera per side. The camera is a low-cost model that offers frame rates up to 125 Hz at QVGA resolution (320 x 240 pixels) using an alternative driver (see \nameref{S1_Text}). For the lighting setup, it is necessary to use a constant light source, such as LEDs. Pulsed light sources, such as fluorescent lamps, may introduce flicker to the video, yielding suboptimal detection results. Our setup was illuminated by an array of infrared IR-LED clusters (840 nm wavelength). The entire structure was enveloped with a highly IR reflective foil with small embossments for light dispersion. The IR LED clusters pointed towards the foil to create a homogeneous ambient lighting and reduce reflections on the glass panes. The built in IR block of the PS3eye cameras had to be removed to make them IR sensitive.

\subsection*{Target features}
The relation of site properties (distance and direction to the feeder) and dance properties (duration and angle) have been recorded via systematic experiments ~\cite{Frisch1965a}. The following equations will be used for an approximate inverse mapping of dance parameters to site properties.

\begin{equation}
	r_R \sim f_d^{-1}\left(d_w\right).
	\label{eq:1}
\end{equation}

\begin{equation}
	\theta_R \sim atan2\left(\sum^{n}_{j=1}{sin\alpha_{wj}},\sum^{n}_{j=1}{cos\alpha_{wj}}\right), n=2k.
    \label{eq:2}
\end{equation}

\begin{equation}
	p_R \sim \frac{d_w}{d_r}.
	\label{eq:3}
\end{equation}

Where $r_R$ (Eq~\ref{eq:1}) is the distance between hive and resource. It is related to the average waggle run duration $d_w$ through the function $f_d$ that approximates the calibration curve ~\cite{Visscher1982}. Here, we use a simple linear mapping and use an empirically determined conversion factor. $\theta_R$ (Eq~\ref{eq:2}) is the angle between resource and solar azimuth (see Fig~\ref{fig1}). It corresponds to the average orientation of the waggle runs with respect to the vertical, with an even number of consecutive runs to avoid errors due to the divergence angle ~\cite{Frisch1965a,Weidenmuller1999,Tanner2010,Landgraf2011a}. The resource's profitability $p_R$ (Eq~\ref{eq:3}) is proportional to the ratio between average waggle run duration $d_w$ and average return run duration $d_r$. Dances for high-quality resources contain shorter return runs than those for less profitable resources located at the same distance, hence yielding a higher $p_R$ value ~\cite{Seeley2000}.

\begin{figure}[!ht]
\centering
\includegraphics[width=1.0\textwidth]{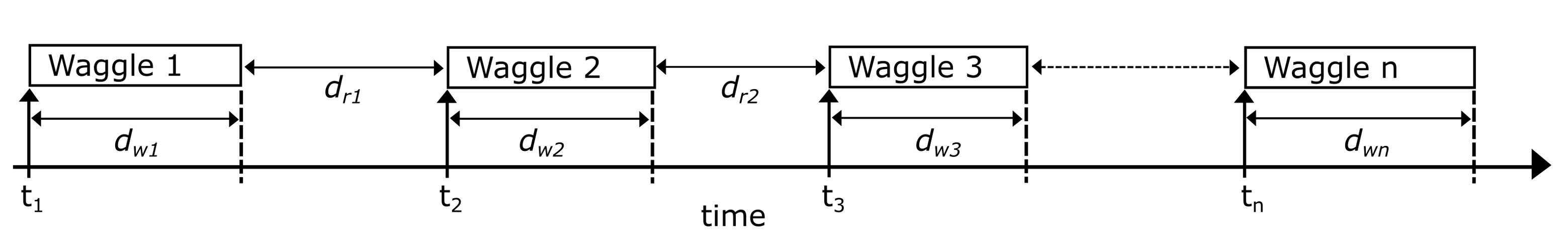}
\caption{{\bf Fundamental parameters.}
Knowing the starting time ($t_x$) and duration ($d_{wx}$) for each waggle run, it is possible to calculate the return run durations as the time gaps between consecutive waggle runs.}
\label{fig3}
\end{figure}

From Eq~\ref{eq:1} to Eq~\ref{eq:3} it follows that to decode the information contained in a communication dance three measurements are required: average waggle run duration $d_w$, average orientation $\alpha_w$, and average return run duration $d_r$. In contrast to some approaches that require tracing the dance path to then analyze it and extract its characteristics ~\cite{Khan2004,Kimura2011a}, we propose an algorithm that directly analyzes video frames to obtain each waggle run's starting timestamp, duration, and angle. In our approach, return run durations are calculated as the time difference between the end of a waggle run and the beginning of the next one Fig~\ref{fig3}.

\subsection*{Software modules}
Our software consists of four modules that are executed in sequence, namely: attention module (\emph{AM}), filter network (\emph{FN}), waggle orientation module (\emph{OM}) and mapping module (\emph{MM}). The AM runs in real-time and stores small subregions of the video containing waggle-like activity. Later, false positives are filtered out using a convolutional neural network \emph{FN}. The \emph{OM} extracts the duration and angle of the waggle runs. Finally, the mapping module (\emph{MM}) clusters waggle runs belonging to the same dances and maps them back to field coordinates. All modules can run offline on video recordings. Long observations, however, require large storage space. Therefore, we propose using the \emph{AM} with a real-time camera input to reduce the amount of stored data drastically. A detailed description of all four modules is given in the following sections.

\subsubsection*{Attention module (\emph{AM})}
The waggle frequency of dancing bees lies within a particular range we call the \emph{waggle band} ~\cite{Landgraf2007a,Gil2010}. Under consistent lighting conditions, the bee body is well discriminable from the background; therefore, the brightness dynamics of each pixel in the image originates from either honey bee activity or sensor noise. Thus, when a pixel intersects with a dancing bee, its intensity time-series is a function of the texture pattern on the bee and her motion dynamics. Indeed, by using a camera with low spatial resolution, bees appear as homogeneous ellipsoid blobs without surface texture. Thus the brightness of pixels that are crossed by waggling bees varies with the periodic waggle motion, and the frequency spectrum of that time series exhibits components in the \emph{waggle band} or harmonics. To detect this general feature, we define a binary classifier here on referred to as \emph{Dot Detector} (\emph{DD}), each pixel position $[i,j]$ is associated to a \emph{DD} $D_{ij}$. The \emph{DD} analyzes the intensity evolution of the pixel within a sliding window of width $b$. For this purpose, the last $b$ intensity values of each pixel are stored in a vector $B$, which at the time $n$ can be described as $B_{ij}^n=\left[v_{ij}^{n-b+1}, v_{ij}^{n-b+2},\ldots,v_{ij}^{n}\right]$, where $v_{ij}^k$ is the intensity value of the pixel $[i,j]$ at time $k$. We calculate a score for each of these time series using a number of sinusoidal basis functions, in principle similar to the Discrete Fourier Transform ~\cite{Cooley1965}:

\begin{equation}
score\left(\bar{B}_{ij}^{n},r\right) = \sum^{b}_{m=1} \left(\left(\bar{B}_{ij}^{n}\left(m\right) \cdot cos\left(2\pi r \frac{m}{s_r}\right)\right)^2 + \left(\bar{B}_{ij}^{n}\left(m\right) \cdot sin\left(2\pi r \frac{m}{s_r}\right)\right)^2\right)
\label{eq:4}
\end{equation}

where $\bar{B}_{ij}^{n}$ is the normalized version of $B_{ij}^{n}$ with $\bar{B}_{ij}^{n} \in \left\{-1,1\right\}$ and $min_{ij}^n=min\left(B_{ij}^n\right)$ and $max_{ij}^n=max\left(B_{ij}^n\right)$, $s_r$ is the video's sample rate (100 Hz), and $r \in \left[10,16\right]$ are the frequencies in the waggle band. If at least one of the frequencies in the \emph{waggle band} scores over a defined threshold $t_h$, $D_{ij}$ is set to 1. After computing the scores, those $D_{ij}$ set to 1 are clustered together following a hierarchical agglomerative clustering (\emph{HAC}) approach ~\cite{Sibson1973}, using as a metric the Euclidean distance between pixels and with a threshold $d_{max1}$ set to half the body length of a honey bee. Clusters formed by less than $c_{min1}$ \emph{DDs} are discarded as noise-induced, and the centroids of the remaining clusters are regarded as positions of potential dancers.

Positions found during the clustering step are then used to detect waggle runs (\emph{WR}). If positions detected in successive frames are located within a maximum distance $d_{max2}$,  defined according to the average waggle forward velocity (see ~\cite{Landgraf2011a}), the positions are considered as belonging to the same \emph{WR}. At each iteration new dancer positions are matched against open \emph{WR} candidates, and either appended to a candidate or used as basis for a new one. A \emph{WR} candidate can remain open up to $g_{max2}$ frames without new detections being added. If no detections could be added it is closed. Only closed WR candidates with a minimum of $c_{min2}$ detections are retrieved as \emph{WRs}. Finally, coordinates of the potential dancer, along with 50 x 50 pixels image snippets of the \emph{WR} sequence are stored to disk.

The operation of the \emph{AM} can be seen as a three layers process summarized in the following points:

\begin{enumerate}
\item Layer 0, for each new frame $I^n$:
\begin{enumerate}
\item Update $DDs$' score vector.
\item Set to 1 $DDs$ with spectrum components in the waggle band above $t_h$
\end{enumerate}

\item Layer 1, detecting potential dancers:
\begin{enumerate}
\item Cluster together $DDs$ potentially activated by the same dancer.
\item Filter out clusters with less than $c_{min1}$ elements.
\item Retrieve clusters' centroids as coordinates for potential dancers.
\end{enumerate}

\item Layer 2, detecting waggle runs:
\begin{enumerate}
\item Create waggle run assumptions by concatenating dancers positions with a maximum Euclidean distance of $d_{max2}$.
\item Assumptions with a minimum of $c_{min2}$ elements are considered as real \emph{WR}.
\end{enumerate}

\end{enumerate}

\subsubsection*{Filtering with convolutional neural network (\emph{FN})}
For long term observations we propose using the \emph{AM} to filter relevant activity from a camera stream in real-time. This significantly reduces the disk space otherwise required to store full sized videos. Depending on the task at hand, it might e.g. be advisable to configure the module to never miss a dance. A higher sensitivity, however, might come with a higher number of false detections. For this use-case, we have trained a convolutional neural network that processes the sequence of 50 x 50 px images to discard non-waggles. The scalar output of the network is then thresholded to predict whether the input sequence contains a waggle dance. The network is a 3D convolutional network whose convolution and pooling layers are extended to the 3rd, i.e. temporal, dimension ~\cite{Ji2013,Tran2014,Srivastava2014b,Klambauer2017}. The network architecture, three convolutional and two fully connected layers, is rather simple but suffices for the filtering tasks (for details refer to \nameref{S1_Fig}).

The network was trained on 8239 manually labeled AM detections from two separate days. During training, subsequences consisting of 128 frames were randomly sampled from the detections for each mini-batch. Detections with less than 128 frames were padded with constant zeros. Twenty percent of the manually labeled data was reserved for validation. To reduce overfitting, the sequences were randomly flipped on the horizontal and vertical axes during training. We used the Adam optimizer ~\cite{Kingma2014} to train the network and achieved an accuracy of 90.07\% on the validation set. This corresponds to a recall of 89.8\% at 95\% precision.

\subsubsection*{Orientation module (\emph{OM})}
\begin{figure}[!ht]
\centering
\includegraphics[width=0.8\textwidth]{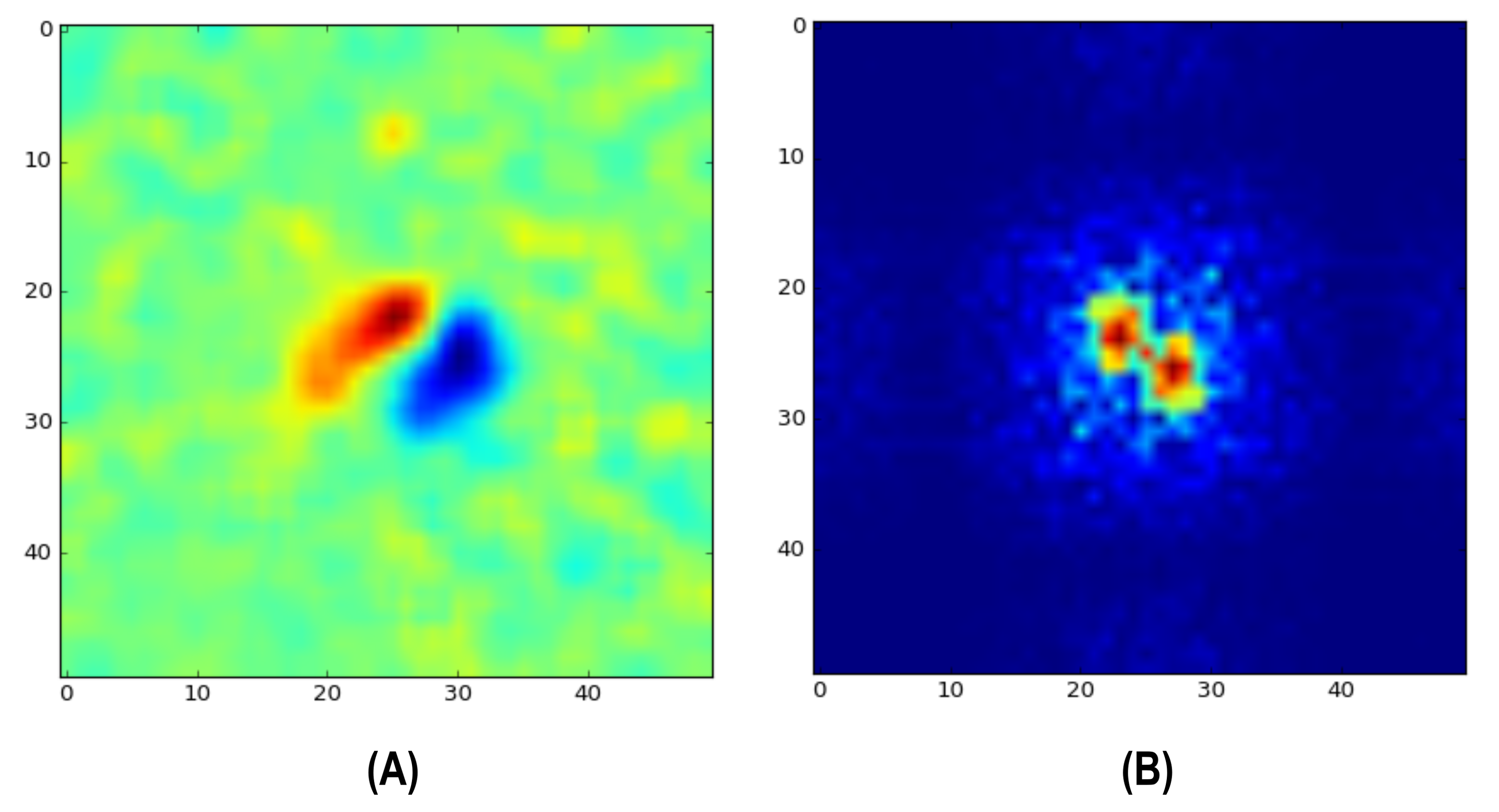}
\caption{{\bf Difference image and its Fourier transformation.}
(A) The image resulting from subtracting consecutive video frames of waggling bees exhibits a characteristic Gabor filter-like pattern. (B) While the peak location varies in image space along with the dancer's position, its representation in the Fourier space is location-independent, showing distinctive peaks at frequencies related to the size and distance of the Gabor-like pattern.}
\label{fig4}
\end{figure}

While the duration of a \emph{WR} is estimated from the number of frames exported by the \emph{AM}, its orientation is computed in a separate processing step here on referred as orientation module (\emph{OM}), usually performed offline to keep computing resources free for detecting waggle runs. Dancing bees move particularly fast during waggle runs, throwing their body from side to side at a frequency of about 13 Hz \cite{Landgraf2011a}. Images resulting from subtracting consecutive video frames of waggling bees exhibit a characteristic pattern similar to a 2D Gabor filter, a positive peak next to a negative peak, whose orientation is aligned with the dancer's body Fig~\ref{fig4}.

\begin{figure}[!ht]
\centering
\includegraphics[width=1.0\textwidth]{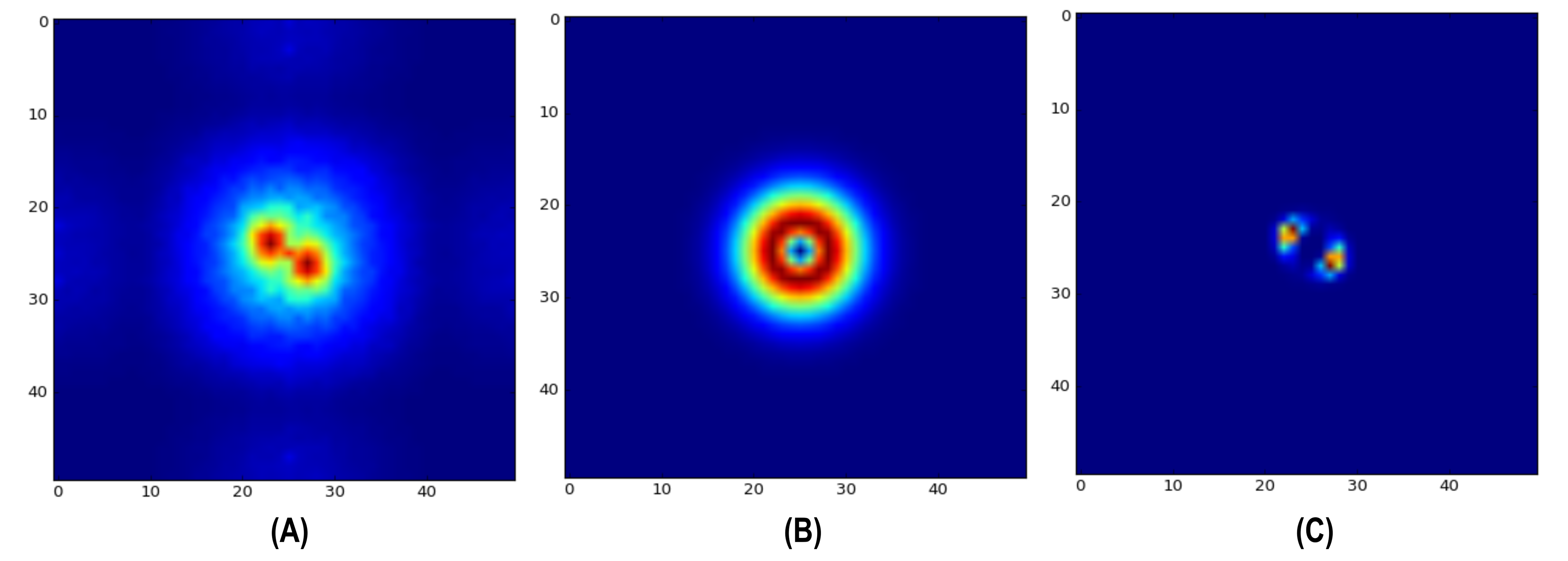}
\caption{{\bf Filtering cumulative sum of difference images in the Fourier space.}
(A) The cumulative sum of the Fourier transformed difference images of a waggle run exhibit a strong pair of maxima in locations orthogonal to the dancer's orientation. (B) A \emph{DoG} kernel of the Mexican hat type, properly adjusted to the waggle band, is used as a bandpass filter. (C) Bandpass filtering the cumulative sums emphasizes values within the frequencies of interest.}
\label{fig5}
\end{figure}

The Fourier transformation of the difference image provides a location-independent representation of the waggling event while preserving information regarding the dancer's orientation (Fig~\ref{fig4}B). We make use of the Fourier slice theorem ~\cite{Bracewell2003}, which states that the Fourier transform of a projection of the original function onto a line at an angle $\alpha$ is just a slice through the Fourier transform at the same angle. Imagine a line orthogonal to the dancer's orientation. If we project the Gabor-like pattern onto this line, we obtain a clear sinusoidal pattern which appears as a strong pair of maxima in the Fourier space at the same angle. Not all difference images in a given image sequence exhibit the Gabor pattern, it only appears when the bee is quickly moving laterally. To get a robust estimate of the waggle orientation, we sum all Fourier transformed difference images Fig~\ref{fig5}A and apply a bandpass filter Fig~\ref{fig5}B to obtain the correct maxima locations Fig~\ref{fig5}C. The bandpass filter is performed in the frequency domain by multiplying with a difference-of-Gaussians \emph{DoG}. The radius of the ring needs to be tuned to the expected frequency of the sinusoidal in the 1D projection of the Gabor pattern. This frequency depends on the frame rate (in our case 100 Hz) and the image resolution (17 px/mm). With the lateral velocity of the bee (we used the descriptive statistics in ~\cite{Landgraf2011a}) one can compute the displacement in pixels (5 to 7 px/frame). Using Eq~\ref{eq:5} we can approximate the value of the expected frequency $k$:

\begin{equation}
k = \frac{I_{size}}{T} = \frac{I_{size}}{2x},
\label{eq:5}
\end{equation}

where $I_{size}$ is the input image size (50 px in this case) and $T$ is the period for the Gabor filter-like pattern or twice the bee's displacement between frames.

The orientation of the waggle run is obtained from the resulting image through Principal Component Analysis (\emph{PCA}). However, the principal direction reflects the direction of the dancer's lateral movements, so it is necessary to add 90° to this direction to obtain the dancer's body axis. This axis represents two possible waggle directions. To disambiguate the alternatives, we process the dot detector positions extracted by the \emph{AM}. Each of these image positions represents the average pixel position in which we found brightness changes in the waggle band. In a typical waggle sequence, these points trace roughly the path of the dancer. We average all \emph{DD} positions of the first 10\% of the waggle sequence and compute all \emph{DD} positions relative to this average. We then search for the maximum values in the histogram of the orientation of all vectors and average their direction for a robust estimate of the main direction of the dot detector sequence. This direction is then used to disambiguate the two possible directions extracted by \emph{PCA}.

\subsubsection*{Mapping module (\emph{MM})}
Waggle dances encode polar coordinates for field locations. To map these coordinates back to the field we implemented a series of steps in what is here on referred as mapping module (\emph{MM}). The \emph{MM} reads the output of the \emph{AM} and \emph{OM}, essentially time, location, duration and orientation of each detected waggle run. Then, waggle runs are clustered following a \emph{HAC} approach, similar to the \emph{AM} (see \nameref{S2_Fig}). In this case, the clustering process is carried out in a three-dimensional data space defined by the axes $X$ and $Y$ of the comb surface and a third axis $T$ of time of occurrence. This way, each \emph{WR} can be represented in the data space by $(x,y,t)$ coordinates based on its comb location and time of occurrence (see \nameref{S3_Fig}A). To maintain coherence between spatial and temporal values, the time of occurrence is represented in one fourth of the seconds relative to the beginning of the day.

A threshold Euclidean distance $d_{max3}$ is defined as a parameter for the clustering process (see \nameref{S3_Fig}B). The value of the threshold is based on the average drift between \emph{WRs} and the average time gap between consecutive \emph{WRs} (we used the data provided in ~\cite{Landgraf2011a}). We only consider clusters with a minimum of 4 waggle runs as actual dances ~\cite{Couvillon2012a}. Then, we use random sample consensus (\emph{RANSAC}) ~\cite{Fischler1981} to find outliers in the distribution of waggle run orientations. Waggle run duration and orientation are then averaged for all inliers and translated to field locations. The mean waggle run duration is translated to meters using a conversion factor, and its orientation is translated to the field with reference to the azimuth at the time of the dance. The duration-to-distance conversion factor was empirically determined by averaging the durations of waggle runs advertising a known feeder (see Discussion for further details).

\section*{Experimental validation and results}

To evaluate the distance decoding accuracy, we ran the AM on a set of video sequences containing a total of 200 WRs. These videos were recorded for a another research question using different hardware (for details refer to \cite{Landgraf2011a}). The duration exported by the AM for each WR was compared to manually labeled ground truth. We found that the AM overestimated WR durations on average by 98 ms, with an SD of 139 ms.

To evaluate the performance of the \emph{OM}, we reviewed the video snippets exported and filtered by the attention module AM and the filter network FN, respectively. Eight coworkers defined the correct waggle run orientation for a set of 200 waggle runs. A custom user interface allowed tracing a line that best fits the dancer's body (see \nameref{S2_Text}). The reference angle for each waggle run was defined as the average of the eight manually extracted angles. The \emph{OM} performed with an average error of -2.92° and a SD of 7.37°, close to the SD of 6.66° observed in the human-generated data (further details in \nameref{S3_Text}).

\begin{figure}[!ht]
\centering
\includegraphics[width=1.0\textwidth]{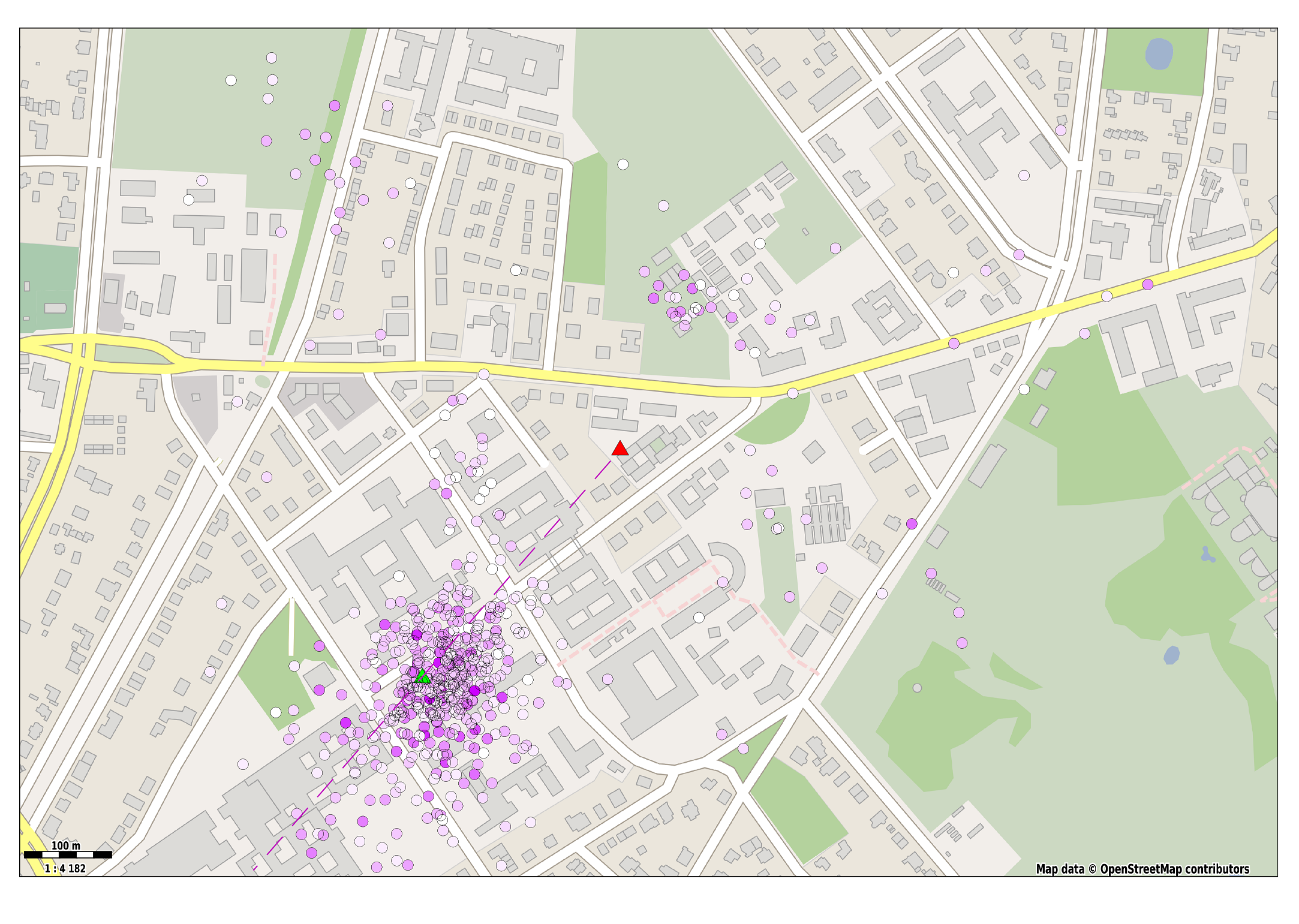}
\caption{{\bf Detected dances mapped back to the field.}
The average duration and orientation of four or more waggle runs per dance were translated to a field coordinate. The spots' color saturation denotes the number of waggle runs in the dance, white corresponding to the dances with four waggle runs and deep purple to the those with the maximum number (17 for this particular data set). A linear mapping was used to convert waggle duration to distance from the hive. The hive and feeder positions are depicted with a red and green triangle, respectively. The dashed line represents the average direction of all dances (Map data copyrighted OpenStreetMap contributors and available from https://www.openstreetmap.org ~\cite{OpenStreetMap}).}
\label{fig6}
\end{figure}

To illustrate the use cases of the automatic decoding of waggle dances we mapped all dances detected by our system during a period of 5 hours. The data was collected from a honey bee colony kept under constant observation during the summer of 2016. A group of foragers from the colony was trained to an artificial feeder placed 342 m southwest of the hive. Fig~\ref{fig6} depicts the distribution of coordinates converted from 571 dances the system detected. The color saturation of each circle encodes the number of waggle runs associated (5.8 WR on average). Since our bees were allowed to forage from other food sources, not all of the detected dances point towards the artificial feeder. However, most of the detected dances cluster around the feeding site. By averaging the direction of all 571 dances we obtained a very precise match with the artificial feeder's direction, with an angular error below 2.35°. If we select only dances in an interval of $\pm$45° around the feeder direction we obtain an angular deviation of 2.33° $\pm$ 11.12° on the dance level and 2.68° $\pm$ 14.37° on the waggle level.

\section*{Discussion}

We have presented the first automatic waggle dance detection and decoding system. It is open source and available for free. It does not require expensive camera hardware and works with standard desktop computers. The system can be used continuously for months, and its accuracy  ($\mu$: -3.3° $\pm$ 5.5°) is close to human performance ($\mu$: 0° $\pm$ 3.7°). 

We investigated the possible error sources and visually inspected waggle runs decoded with large error. Before the outlier detection step we find several waggle orientations decoded with an 180° error. The orientation of 9\% of the waggle runs in our test set were incorrectly flipped. This flip error is corrected in the mapping module by clustering waggle runs to dances and removing angular outliers with \emph{RANSAC}. Once the outliers have been removed, our system performs with an average error of -2.0° $\pm$ 6.1° on the waggle run level. In this process, we discard dances with no strong mode in its waggle orientation distribution and, theoretically, it is therefore possible that a few undiscarded dances contain only flipped waggle runs.

In most of these examples the dancing bee was partly occluded or wagged her body for only short durations, i.e. there is almost no forward motion visible. The forward motion is the central feature in the orientation reader module used to disambiguate the direction. Recognizing anatomical features of the dancing bee, such as the head or abdomen, could help reducing this common error. With falling costs for better camera and computing hardware in mind, we, however, think that, although some bees just don't move forward while wagging, using a higher spatial resolution will likely resolve most of the detection and decoding errors we have described.

The error of the distance decoding could only be assessed for the offline mode of operation, i.e. on prerecorded videos. We found a systematic error that can be ignored with a properly calibrated system. The standard deviation of the waggle duration error was found to be 139 ms. This result highly depends on the choice of the threshold value $t_h$. We determined the default value of $t_h$ empirically for optimal waggle detection accuracy. We did not explicitly optimize this parameter for more accurate distance decoding.

Bees encode accumulated optical flow rather than metric distances in their waggle runs. Neither the internal calibration, nor the external factors that influence a bee's perception of optical flow were assessable. Hence we calibrated our system with the collective calibration of the very colony under observation: We extracted all waggle runs signaling the location of our artificial feeder ($\pm 10 ^{\circ}$) and averaged all waggle durations ($\mu$ = 582.79 ms $\pm$ 196.10 ms). It is unlikely that the set of waggle runs could have contained waggle runs signalling other feeders since natural food sources were scarce in that time of year. This notion is supported by a coefficient of variation of $\approx 0.34$, consistent with the value observed by Landgraf et al. in ~\cite{Landgraf2011a}.

Given the high accuracy of the method, why do the projected dances in Fig~\ref{fig6} exhibit such a large spread? We inspected random samples and found that not all of the dances advertised our feeding site. Thus, the vector endpoint distribution shows smaller clusters that likely represent natural food sources. The variation of the dance points around the feeding site is correctly reproduced with a large part of the variation originating from the animals themselves. This imprecision is well-know and caused by waggle runs missing the correct direction, with alternating sign. The difference between consecutive waggle runs, or divergence, is surprisingly large and has been studied previously ~\cite{Towne1988,Weidenmuller1999,Tanner2006}. The divergence correlates negatively with distance to the advertised goal, i.e. it is largest for short distances. In a previous work ~\cite{Landgraf2011a}, we analyzed dances to a 215 m distant food source and tracked the motion of all dancers, corrected the tracker manually whenever necessary, and computed the distribution of waggle directions of over 1000 waggle runs. We found that although the average waggle orientation was surprisingly accurate, it was astonishingly imprecise ($\mu = -0.03^{\circ}, \sigma = 28.06^{\circ}$). A similar result was obtained in the present study. The average direction of all dances matches the direction to the feeding station closely ($\Delta = 2.68^{\circ}$) with a standard deviation of $\sigma = 14.37$. The spread of dance endpoints, however, is smaller due to the integration of at least four waggle runs ($\sigma = 11.12^{\circ}$. Using single waggles or short dances to pinpoint foraging locations of individuals can therefore unlikely be accurate and it is clear that the number of dances to be mapped needs to be tuned to the given scientific context and environmental structure. 
We excluded waggle detections shorter than 200 ms, a timespan that would contain less than three body oscillations. Remarkably, bees shake their body in short pulses quite frequently even in non-dance behaviors and hence, the number of false positive detections increases with lower thresholds. Round dances, a dance type performed to advertise nearby resources ~\cite{Frisch1965a}, may also contain short waggle portions. Although our system may be able to detect these, the waggle oscillation is an unreliable feature for round dances. We therefore explicitly focus on waggle dances. Note that the sharp cutoff of dance detections close to the hive in Fig~\ref{fig6} stems from discarding short waggle runs.

The presented system is unique in its approach and capabilities. There are, however, still some features missing that might be added in the future. The mapping module, e.g., does not yet extract and visualize the profitability of a food source. One could, e.g. calculate the return run duration in the clustering step and use a color coding scheme to encode this information into the map. 
Bee dances also exhibit a systematic angular error that depends on the waggle orientation on the comb ("Restmissweisung", ~\cite{Frisch1965a}). To improve mapping accuracy, we plan to add a correction step to the mapping module.

The proposed system consists of multiple modules executable as command line programs. Although well documented, this might seem impractical or even obfuscating to the end-user. We are thus developing a graphical user interface to be published in the near future. We currently investigate whether a deep convolutional network is able to extract the relevant image features. If successful, this would enable us to merge the filter network module and the orientation reader, therefore reducing the system's complexity for the user. For the future, we envision an entirely neural system for all the described stages. We also think the solution could be ported to mobile devices. This would enable users an easier setup. Dance orientations could be corrected by reading the direction of gravity directly from the built-in accelerometer. 
We would like to encourage biologists to use our system and report issues that they face in experiments. Interested software developers are invited to help improving existing features or implementing new ones.

\section*{Supporting Information}

\paragraph*{S1 Text.}
\label{S1_Text}
{\bf Specifications of the recording setup used during reported experiments.} This document contains further information on the recording setup, with an emphasis on technical details.

\paragraph*{S2 Text.}
\label{S2_Text}
{\bf Further details on the software modules.} This document contains diagrams and detailed information on the functioning of the software modules.

\paragraph*{S3 Text.}
\label{S3_Text}
{\bf Error distributions at multiple levels of analysis.} This document provides additional results supporting the case of study presented in the experimental validation and results section.

\paragraph*{S1 Fig.}
\label{S1_Fig}
{\bf A convolutional neural network was used to filter the detections of the AM.} The raw sequences of images are processed by two stacked 3D convolution layers with SELU nonlinearities. The outputs of the second convolutional layer are flattened using average pooling on all three dimensions. A final fully connected layer with a sigmoid nonlinearity computes the probability of the sequence being a dance or not. Dropout is applied after the average pooling operation to reduce overfitting.

\paragraph*{S2 Fig.}
\label{S2_Fig}
{\bf Dendrogram representing the hierarchical agglomerative clustering of 200 waggle runs.} The dendrogram is a graphic representation of the clustering process. Each observation starts it its own cluster, at each iteration the two clusters closer to each other are merged, this process is performed recursively till only one cluster remains. We set a threshold distance for clusters to be merged, all clusters generated to the point this threshold is reached are regarded as dances and their constituent elements as their waggle runs.

\paragraph*{S3 Fig.}
\label{S3_Fig}
{\bf Representation of \emph{WRs} in the data space XYT.} (A) Representation of a set of 200 \emph{WRs} in the XYT data space, where values in the axes $X$ and $Y$ are defined by their comb location, and in axis $T$ by their time of occurrence. (B) \emph{WRs} within a maximum Euclidean distance of $d_{max3}$ are clustered together and regarded as dances.


\section*{Acknowledgments}
The authors thank M. Halilovic, F. Lojewski, A. Rau and S. Witte for their contribution and support conducting the experiments and in the development of this system. This work is funded in part by the German Academic Exchange Service (DAAD).

\bibliography{library}

\bibliographystyle{abbrv}

\end{document}